\documentclass[sigconf,nonacm]{acmart}
\setcopyright{none}
\usepackage{amsmath}
\usepackage{graphicx}
\usepackage{multirow}
\usepackage{placeins}
\usepackage{hyperref}
\AtBeginDocument{}
\begin{document}
\title{Receiver-Conditioned Clarification for Multi-Agent Food Assistance}
\author{Kaiwen Shi}
\email{kshi3@nd.edu}
\affiliation{\institution{University of Notre Dame}\city{Notre Dame}\state{Indiana}\country{USA}}
\author{Zheyuan Zhang}
\email{zzhang42@nd.edu}
\authornote{Corresponding authors.}
\affiliation{\institution{University of Notre Dame}\city{Notre Dame}\state{Indiana}\country{USA}}
\author{Changzhi Liu}
\affiliation{\institution{University of Notre Dame}\city{Notre Dame}\state{Indiana}\country{USA}}
\author{Han Bao}
\affiliation{\institution{University of Notre Dame}\city{Notre Dame}\state{Indiana}\country{USA}}
\author{Yanfang Ye}
\email{yye7@nd.edu}
\affiliation{\institution{University of Notre Dame}\city{Notre Dame}\state{Indiana}\country{USA}}
\renewcommand{\shortauthors}{Shi et al.}
\begin{abstract}
Food insecurity creates a pressing need for guidance that helps households find usable food resources and assess food choices in light of their circumstances. Such guidance must connect eligibility requirements, practical access constraints, and dietary needs rather than return isolated listings or generic nutritional advice. Multi-agent food assistants can coordinate these tasks, with uncertainty indicating when clarification or verification is needed before a recommendation. However, uncertainty can be lost across agent handoffs, allowing tentative interpretations to propagate as facts and obscuring opportunities for correction. Moreover, confidence that a service requirement has been interpreted correctly is not confidence that a recommendation satisfies the user's needs; local confidence cannot simply be reused as downstream reliability. We introduce an inference-time, receiver-conditioned clarification framework that preserves shared premise identities and replays consistent interpretations across dependent agents. Each receiver exposes admissible decisions and remaining evidence obligations, so question relevance reflects both decision changes and missing support. Information questions and user-confirmable fallback adjustments are evaluated together with their anticipated follow-up interactions under shared budgets. This structure connects clarification to jointly supported plans for resource navigation, nutritional reasoning, and supply-side planning. Our code is available \href{https://github.com/shiiiikw/Receiver-Conditioned-Clarification-for-Multi-Agent-Food-Assistance}{here}.
\end{abstract}
\maketitle
\section{Introduction}
\label{sec:introduction}

\paragraph{Relevance to the Web and the track.}
Conversational access to Web service directories hinges on resolving user constraints before any resource is recommended. This paper shows how a multi-agent assistant preserves uncertain premises across agents and selects the clarification that most improves its downstream decisions under an interaction budget. The work falls squarely within interactive information access, domain-specific recommendation, and agentic search in the \emph{Search, Recommendation, and Retrieval-Augmented AI} track.

\textit{Food insecurity} is defined as limited or uncertain availability of nutritionally adequate and safe food, or limited or uncertain ability to acquire acceptable food through socially acceptable means~\citep{usdaFoodSecurityDefinition}. It remains a pressing societal problem in the United States: the U.S. Department of Agriculture (USDA) estimates that 13.7\% of households (18.3 million) experienced food insecurity at some point in 2024~\citep{usda2026statistics}. Meeting this need requires more than food supply: people must determine where and how to obtain usable food, while providers must decide what to offer and where to reach unmet demand. A food assistant must connect supply with those in need and support both complementary decisions.

On the demand side, nearby resource lists are insufficient: people experiencing food insecurity, including those without stable housing, need resource discovery linked to personal suitability and feasible access plans. Eligibility, documentation, opening hours, and travel constraints determine usability~\citep{li2026hoosierhelp}, while food must meet individual nutritional needs~\citep{zhang2025ngqa}. On the supply side, food banks, pantries, and community meal programs must decide what to provide and where, aligning provision with community preferences and dietary needs~\citep{morello2021foodbanks} and balancing accessibility, equitable coverage, and limited operational capacity in distribution~\citep{stauffer2022equitable}. Provider-facing assistance must therefore compare food and site options against unmet demand and operational feasibility, beyond enumerating candidates.

\begin{figure*}[t]
\centering
\includegraphics[width=\textwidth]{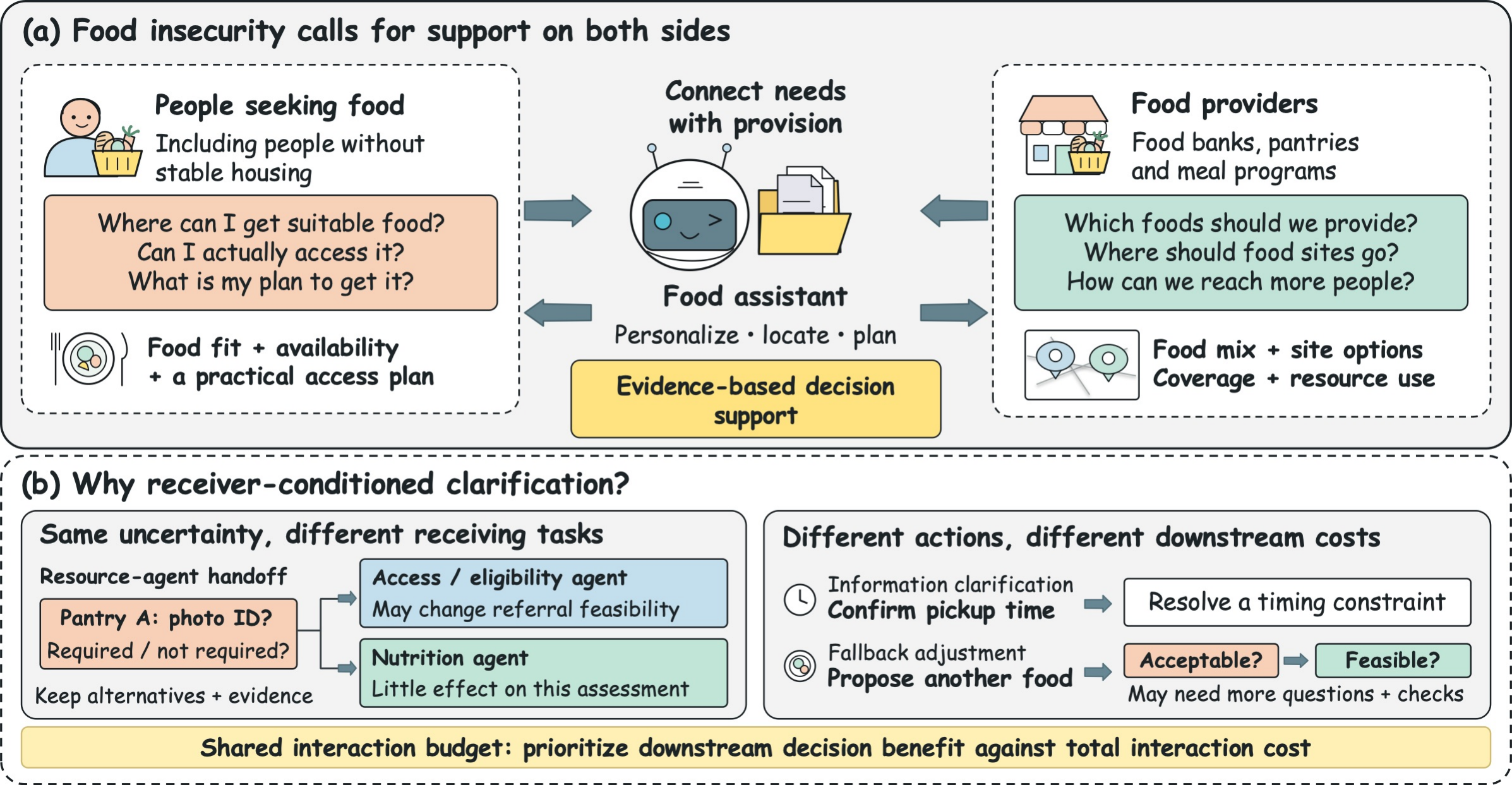}
\caption{
Food assistant motivation. (a) Linking food-access and dietary needs to provision and site planning. (b) Clarification depends on how requirements affect each receiving agent's decision and whether fallback acceptance needs follow-up checks.
}
\Description{Two panels motivate the assistant. The first connects demand-side food access and nutritional needs with supply-side food provision and distribution planning. The second shows clarification guided by the receiving decision and its dependencies, with question benefits balanced against interaction cost under a shared budget.}
\label{fig:food-assistant-motivation}
\end{figure*}

Therefore, effective food assistance must connect individual needs and access constraints with providers' supply and distribution capacity. To support this coordination, we present a conversational food assistant integrating demand-side resource navigation with supply-side provision and distribution planning; it is deployed \href{https://food-assistant-review.vercel.app}{here}. These functions require heterogeneous evidence and distinct but interdependent eligibility, nutritional, and planning judgments with different constraints and verification needs, making it difficult for a single agent to isolate task-specific reasoning and verify intermediate decisions. We therefore build a role-specialized multi-agent system (MAS) that separates task-specific reasoning and verification and coordinates findings into actionable plans.

Role specialization supports distinct judgments but complicates coordination: downstream decisions rely on potentially uncertain intermediate information~\citep{shiuncertainty, ye2025llms4all}. Since users and providers do not fully specify their needs upfront, the assistant must identify which uncertainties to clarify before they affect subsequent decisions, while limiting unnecessary interaction~\citep{bao2026drift,bao2026not}. Value-of-information (VOI) methods provide a decision-theoretic foundation by weighing clarification's expected benefit against communication cost under a dialogue budget~\citep{dong2026voi}. However, existing formulations primarily consider a single decision-maker, whereas MAS decisions depend on shared, potentially uncertain premises. This distinction introduces three challenges. First, a shared uncertain premise may affect receivers with different evidence requirements, making clarification value depend on their decisions. Second, handoffs may lose alternative interpretations, preventing receivers from evaluating how clarification would change their decisions~\citep{shi2026confidence}. Third, multiple agents' reliance on the same upstream evidence may resemble independent confirmation and inflate confidence~\citep{liu2026propuq,huang2026cage}. Together, these challenges show that uncertainty propagation or calibration alone is insufficient: clarification must account for shared uncertainty's effects on downstream decisions and their evidence dependencies.

To address these challenges, we propose a receiver-conditioned clarification framework that evaluates questions by their effects on downstream decisions. We preserve alternative interpretations and evidence provenance across handoffs so receivers can reason over unresolved premises rather than collapsed conclusions. During receiver replay, consistent interpretations of shared premises ensure that affected receivers evaluate dependent decisions under the same hypothesis; each receiver produces a \emph{decision signature} recording admissible decisions and outstanding evidence obligations, enabling the controller to identify uncertainties affecting the collective plan. Beyond information gathering, the framework considers user-confirmable fallback adjustments when constraints prevent feasible plans, including the additional interactions needed to implement them. Both clarification types share an interaction budget, allowing the controller to prioritize questions by their expected contribution to an actionable plan.

We integrate this framework into our conversational food assistant to coordinate resource navigation, nutritional reasoning, and provider-side planning, evaluating the system across these complementary tasks on HoosierHelp, NGQA, and FoodGo, respectively. Our approach outperforms the compared baselines across all three tasks, while controlled ablations examine the clarification mechanisms' contributions.

\noindent\textbf{Our contributions are summarized as follows.}
\begin{itemize}
    \item We build and deploy a conversational food assistant that, to our knowledge, is the first to span both the demand and supply sides of food assistance, supporting personalized resource navigation, nutritional reasoning, and evidence-grounded food provision and distribution planning in one system.

    \item We develop a receiver-conditioned clarification framework that extends single-decision VOI to dependent receiver decisions. It enforces consistent shared-premise assignment during replay, records decision signatures that retain admissible decisions and outstanding evidence obligations, and jointly values information clarification, fallback acceptance, and their anticipated follow-up interactions under one budget, with responses driving revision and replanning.

    \item We evaluate the assistant on HoosierHelp, NGQA, and a FoodGo-based task, covering resource navigation, nutritional reasoning, and transfer-site selection. Our approach outperforms all compared baselines in every setting, and atomic-question and component ablations confirm that each clarification mechanism contributes to the gains.
\end{itemize}

\section{Related Work}
\label{sec:related-work}

\paragraph{Food assistance and personalized nutritional reasoning.}
HoosierHelp~\citep{li2026hoosierhelp} evaluates conversational social-service navigation on real service records with simulated users, covering composite needs, constraint fallback, and non-ideal interactions.
NGQA~\citep{zhang2025ngqa} evaluates personalized nutritional reasoning over user, food, and nutrient information.
On the provider side, the FoodGo dataset~\citep{rubinstein2020foodgo_data,rubinstein2020foodgo_report} supports food-rescue pickup-and-delivery scheduling.
TRACE~\citep{hasan2026trace} is the closest system: a retrieval-first pipeline for public-service directories that parses a query into constraints and enforces them during retrieval.
All of these take the user's constraints as given.
None asks which constraint is still unknown, which decision it would change, or whether asking is worth the interaction; that is the subject of this paper.
We use the three benchmarks as evaluation foundations; TRACE-style retrieval is a drop-in replacement for our retriever, not an alternative to the controller that precedes it.

\paragraph{Uncertainty and dependencies in multi-agent systems.}
PropUQ-MAS~\citep{liu2026propuq} propagates local uncertainty with risk inherited over a communication graph; CAGE-Cal~\citep{huang2026cage} calibrates confidence against a counterfactual non-communicating graph.
Both establish that inter-agent agreement is not independent confirmation, and we build on that premise.
But a calibrated number on a message does not tell the consuming task whether the uncertainty matters to it or whether resolving it would change what it does: propagation and calibration are inputs to clarification, not a policy.
From the execution side, VeriMAP~\citep{xu2026verimap} plans subtasks as a DAG with structured I/O and planner-generated verification functions, and ToCoMAS~\citep{zhao2026tocomas} derives agent responsibilities, handoffs, and memory boundaries from the task graph.
Both decide \emph{how} dependent agents execute and assume the premise being passed is already resolved; neither can choose which unresolved premise to ask about or value that question against the decisions downstream.
Our framework fills exactly that gap and is complementary to all four: decision signatures and the dependency footprint can be read off any DAG-structured plan.

\paragraph{Clarification and information-value control.}
MAC~\citep{acikgoz2025mac} coordinates clarification between a supervisor and domain experts using a taxonomy of user ambiguities.
\citet{dong2026voi} instantiate inference-time VOI through hypothetical answers, posterior decision utility, communication costs, and a clarification budget.
\citet{zhang2026control} separate failure prediction from intervention value and learn action-sensitive oversight from alternative continuations.
All three value a question against a single decision-maker's posterior, and that is where they stop short in a multi-agent system: one premise is consumed by several receivers needing different evidence, alternatives are collapsed at handoffs, and a fallback is worth nothing until the follow-ups that make it usable are counted.
We keep their decision-theoretic core and change what is valued: dependent receiver decisions replayed under a shared premise assignment, decision signatures that retain admissible decisions and outstanding obligations, and fallback acceptance valued with its bounded continuation, all under one budget.
No prior method evaluates clarification at this level; doing so is the methodological contribution of this paper, and the ablation in Section~\ref{sec:components} shows it carries the gains.
\section{Methodology}
\label{sec:methodology}

\begin{figure*}[t]
    \centering
    \includegraphics[width=\textwidth]{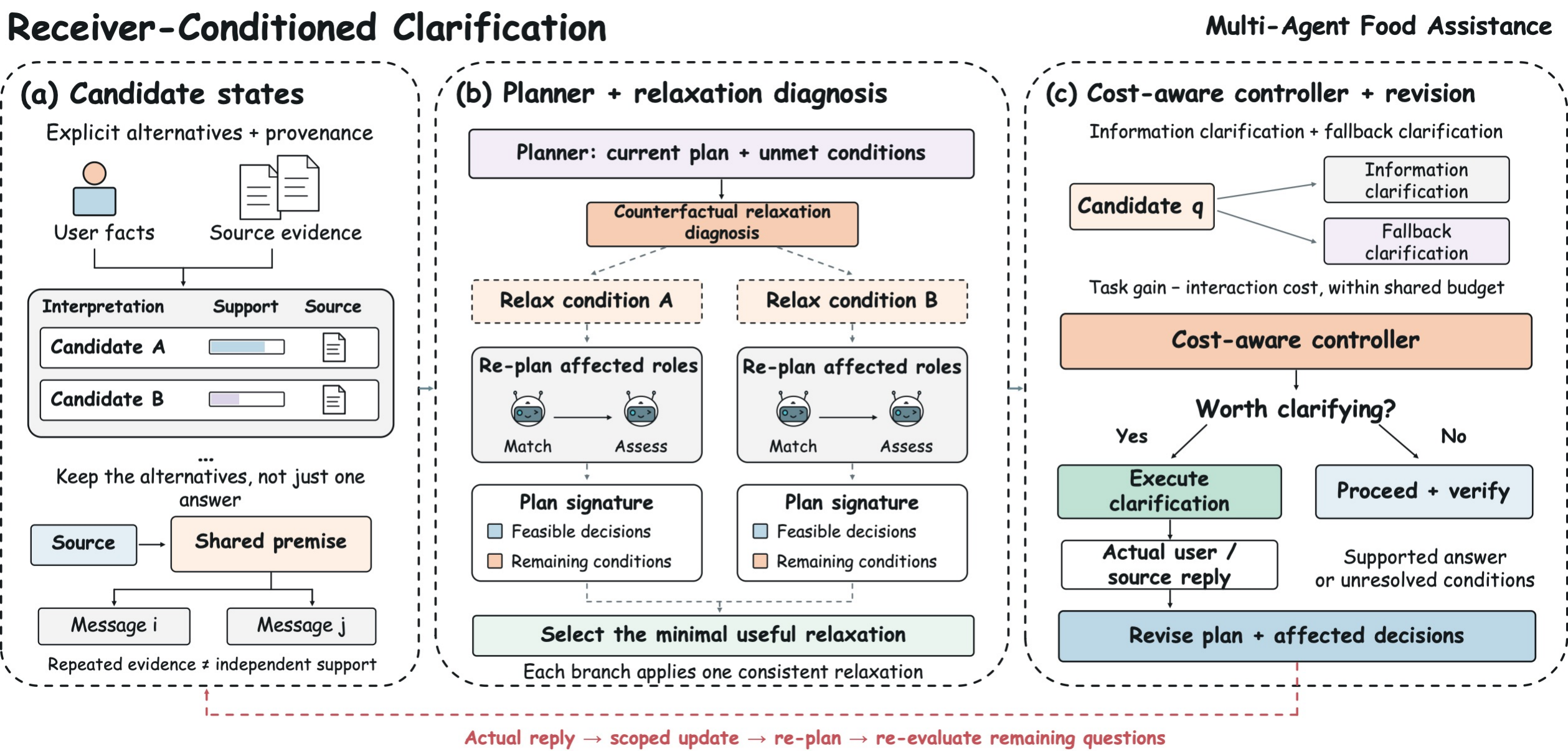}
    \caption{
        Receiver-conditioned clarification framework. (a) Candidate states retain alternatives, evidence provenance, and shared premises. (b) The planner links these premises to receiving decisions and unmet evidence obligations, and diagnoses minimal useful fallback adjustments for blocked plans. (c) The controller values information and fallback questions together with their follow-up costs; actual replies update evidence or accepted preferences and trigger replanning.
    }
    \Description{A clarification pipeline links evidence-grounded candidate interpretations to downstream receiving decisions. It generates decision-relevant questions, estimates their complete-task benefits and interaction costs, and selects an action. When constraints block a plan, counterfactual relaxation diagnosis proposes user-confirmable fallback adjustments. User responses revise the evidence and trigger replanning.}
    \label{fig:clarification-framework}
\end{figure*}

Our framework operates entirely at inference time with frozen language models. Agents preserve candidate interpretations and their dependencies; a planner connects unresolved premises to receiving decisions; and a cost-aware controller selects clarification based on complete-task consequences. When constraints block a plan, counterfactual relaxation diagnosis identifies user-confirmable adjustments that could recover it. Actual responses then revise the evidence and trigger replanning.

\subsection{Problem Formulation}
\label{sec:problem}

In a role-specialized multi-agent workflow, agents exchange intermediate artifacts that become premises for subsequent decisions. The same uncertain premise may affect several receivers differently, making its clarification value depend on the tasks that consume it. We represent these dependencies with a time-unrolled graph $\mathcal{G}_t$. Nodes represent agent invocations, evidence events, and scoped premises, while edges record evidential support and downstream consumption. Each premise identifies its subject, condition, and applicability scope.

At interaction step $t$, the public state is
\begin{equation}
S_t =
(H_t,E_t,\mathcal{M}_t,\mathcal{G}_t,B_t,I_t),
\label{eq:method-state}
\end{equation}
where $H_t$ is the interaction history, $E_t$ is observed evidence, $\mathcal{M}_t$ contains candidate messages, and $B_t$ and $I_t$ are the remaining shared interaction and inference budgets. Active user needs and constraints are extracted from $H_t$ and $E_t$. The controller chooses whether to proceed, clarify an unresolved premise, or request acceptance of a preference adjustment, aiming to complete the task with supported decisions and limited interaction.

\subsection{Evidence-Grounded Candidate States}
\label{sec:candidate-states}

\paragraph{Preserving alternatives and their support.}
A single committed artifact can conceal alternatives that matter to a downstream decision. Each agent therefore generates a bounded set of candidate artifacts using the frozen backbone. A role-specific canonicalizer groups equivalent proposals while retaining the greedy artifact and alternative interpretations. Agent $i$ communicates
\begin{equation}
M_i =
\left\{
(a_{ik},w_{ik},P_{ik},F_{ik})
\right\}_{k},
\label{eq:method-message}
\end{equation}
where $k$ indexes retained candidates, $a_{ik}$ is a candidate artifact, $w_{ik}$ is its empirical proposal support, $P_{ik}$ records its sources and consumed message versions, and $F_{ik}$ contains its scoped assumptions and unresolved requirements. These premise-linked alternatives let receivers assess how clarification changes their decisions.

\paragraph{Shared-premise dependencies.}
The graph links each candidate to its premises and each premise to its originating evidence events. Shared premises retain a common identity across handoffs, and forwarding a claim adds a consumer relationship while preserving its evidence origin. This makes repeated uses of the same information traceable and allows one clarified premise to update all relevant receivers.

Let $\Omega_t$ contain joint interpretations compatible with the observed evidence. Each $\omega\in\Omega_t$ assigns one value $\omega(z)$ to an unresolved premise $z$, which is reused by every dependent receiver during replay. Consequently, hypothetical replies are evaluated through coherent joint states, preventing different agents from constructing a seemingly feasible plan using incompatible assumptions about the same condition.

\subsection{Receiver-Conditioned Decision Relevance}
\label{sec:receiver-relevance}

\paragraph{Planning and decision signatures.}
The planner constructs candidate plans for active needs and identifies the conditions needed to support them. For receiver $j$ and joint interpretation $\omega$, it returns
\begin{equation}
\sigma_j(\omega;S_t)
=
\left(
\mathcal{A}^{\mathrm{adm}}_j(\omega),
\mathcal{O}_j(\omega)
\right),
\label{eq:method-signature}
\end{equation}
where $\mathcal{A}^{\mathrm{adm}}_j$ contains decisions not ruled out by the instantiated conditions, and $\mathcal{O}_j$ records the remaining obligations required to support those decisions. The planner generates information questions from unresolved premises affecting either component.

Interpretations yielding the same supported outcome are immaterial to that receiving decision. Conversely, a decision may remain unchanged while requiring missing evidence. Including $\mathcal{O}_j$ retains such questions instead of judging relevance solely by whether the preferred decision changes.

Each question $q$ has a dependency footprint
\begin{equation}
\mathcal{R}_t(q)
=
\{j\mid v(q)\leadsto j\text{ in }\mathcal{G}_t\},
\label{eq:method-footprint}
\end{equation}
where $v(q)$ is the targeted premise or preference node and $\leadsto$ denotes a directed dependency path. Replaying these receivers reveals how alternative replies affect their decisions and support obligations. Questions targeting the same scoped premise can therefore be shared across receivers, avoiding repeated requests while retaining their combined downstream effects.

\paragraph{Counterfactual relaxation diagnosis.}
When an identified constraint blocks a plan, asking for additional facts may not resolve the incompatibility. The planner instead evaluates eligible user-confirmable preference adjustments. Each adjustment is applied consistently across affected receivers while holding evidence and external requirements fixed. Replanning returns recovered candidate plans and their remaining obligations.

Let $\mathcal{D}_t$ contain the evaluated adjustments, let $r_t(\delta)=1$ indicate that adjustment $\delta$ provides the required recovery, and let $\delta'\prec\delta$ denote a strictly smaller adjustment to the original preferences. The retained alternatives are
\begin{equation}
\mathcal{D}_t^{\min}
=\operatorname{Min}_{\prec}
\{\delta\in\mathcal{D}_t\mid r_t(\delta)=1\},
\label{eq:method-relaxation}
\end{equation}
where $\operatorname{Min}_{\prec}$ selects minimal elements under $\prec$. Retaining minimal useful adjustments avoids requesting larger changes when a smaller evaluated alternative provides the required recovery. The planner converts these alternatives into questions requesting user acceptance and retains any follow-up obligations for subsequent value evaluation.

Together, the generated questions form
\begin{equation}
\mathcal{Q}_t
=
\mathcal{Q}_t^{\mathrm{info}}
\cup
\mathcal{Q}_t^{\mathrm{fallback}}.
\label{eq:method-question-types}
\end{equation}
Here, $\mathcal{Q}_t^{\mathrm{info}}$ resolves missing or ambiguous information, and $\mathcal{Q}_t^{\mathrm{fallback}}$ requests acceptance of a planner-supported adjustment.

\subsection{Dependency-Aware Budgeted VOI}
\label{sec:budgeted-voi}

\paragraph{Complete-task consequence evaluation.}
For each question, the controller simulates possible replies, updates the targeted premise once per branch, and replays the affected receivers. An acceptance branch activates the proposed adjustment. The resulting outcome is scored against the active needs and their evidence requirements as one complete task. This captures the benefit of resolving a shared premise without counting the same improvement repeatedly across agents, while exposing consequences that help one decision but conflict with another.

A question is evaluated together with a bounded, response-contingent continuation that can ask follow-up questions or resume retrieval and repair. Evaluating the continuation captures complementary questions and the additional interactions needed to make an accepted fallback usable, rather than rewarding an intermediate change that leaves the plan blocked.

Let $\pi_{t,q}$ denote the evaluated continuation beginning with $q$, $\tau$ a simulated trajectory, and $S_\tau$ its terminal state. Let $U(S_\tau)$ measure supported task completion and $\widehat{U}_t^{\mathrm{proceed}}$ be the estimated utility of proceeding without further clarification. Then
\begin{equation}
\begin{aligned}
\widehat{\Delta U}_t(q;B_t,I_t)
&=\widehat{\mathbb{E}}_{\tau\sim\pi_{t,q}}[U(S_\tau)]
-\widehat{U}_t^{\mathrm{proceed}},\\
\widehat{C}_t(q;B_t,I_t)
&=\widehat{\mathbb{E}}_{\tau\sim\pi_{t,q}}
\left[\sum_{\ell\in\mathcal{L}(\tau)}c(q_\ell,o_\ell)\right],
\end{aligned}
\label{eq:method-voi-estimates}
\end{equation}
where $\widehat{\mathbb{E}}$ denotes the replay-based expectation estimate, $\mathcal{L}(\tau)$ indexes clarification interactions along $\tau$, and $c(q_\ell,o_\ell)$ is the interaction cost of question $q_\ell$ and reply $o_\ell$, expressed on the utility scale. Thus, $\widehat{\Delta U}_t$ measures improvement over proceeding now, while $\widehat{C}_t$ includes both the initial question and anticipated follow-up costs.

\paragraph{Question selection.}
The controller selects
\begin{equation}
\begin{aligned}
q_t^*
=
\operatorname*{arg\,max}_{q\in\mathcal{Q}_t^{\mathrm{feas}}}
\bigl[
&\widehat{\Delta U}_t(q;B_t,I_t)\\
&-\widehat{C}_t(q;B_t,I_t)
\bigr],
\end{aligned}
\label{eq:method-cost-aware-control}
\end{equation}
where $\mathcal{Q}_t^{\mathrm{feas}}$ contains questions whose evaluated continuations fit the remaining shared interaction and inference budgets. Accounting for continuation feasibility prevents the controller from initiating an apparently inexpensive question while ignoring the resources needed to complete its anticipated follow-ups.

The selected question is issued when its estimated net value is positive; otherwise, the assistant proceeds with verification and repair. Only the first question of the continuation is executed. Its actual reply determines the revised state and the next value comparison.

\paragraph{Selection error bound.}
Let $\mathcal{C}_t=\mathcal{Q}_t^{\mathrm{feas}}\cup\{\mathrm{proceed}\}$, and let $V_t(a)$ and $\widehat{V}_t(a)$ denote the expected and estimated net values of action $a$ for its evaluated continuation. Proceeding has value zero, while a question has estimated value $\widehat{\Delta U}_t-\widehat{C}_t$. Let $a_t^\star=\arg\max_{a\in\mathcal{C}_t}V_t(a)$ and $\widehat{a}_t=\arg\max_{a\in\mathcal{C}_t}\widehat{V}_t(a)$. If the uniform estimation error is at most $\epsilon_t$, then
\begin{equation}
\begin{gathered}
\max_{a\in\mathcal{C}_t}
|\widehat{V}_t(a)-V_t(a)|\leq\epsilon_t\\
\Longrightarrow\quad
V_t(a_t^\star)-V_t(\widehat{a}_t)\leq2\epsilon_t.
\end{gathered}
\label{eq:method-selection-bound}
\end{equation}
\emph{Proof.} Estimated-value maximization gives
$\widehat{V}_t(\widehat{a}_t)\geq\widehat{V}_t(a_t^\star)$.
Applying the error bound to both actions yields
\begin{equation*}
\begin{aligned}
V_t(a_t^\star)
&\leq\widehat{V}_t(a_t^\star)+\epsilon_t\\
&\leq\widehat{V}_t(\widehat{a}_t)+\epsilon_t\\
&\leq V_t(\widehat{a}_t)+2\epsilon_t.
\end{aligned}
\end{equation*}

\subsection{Evidence Revision and Closed-Loop Control}
\label{sec:evidence-revision}

\paragraph{Scoped response integration.}
An actual reply is parsed into an update with a subject, condition, applicability scope, and provenance. Information responses revise the relevant evidence, while explicitly accepted fallback responses update the specified preference. The resulting state is
\begin{equation}
S_{t+1}
=
\operatorname{Revise}(S_t,q_t,o_t),
\label{eq:method-revision}
\end{equation}
where $q_t$ is the issued question and $o_t$ is the actual response. Scope determines which premises are revised and which receivers can consume the update. This lets a shared fact benefit all applicable tasks while preserving requirements unrelated to the response.

\paragraph{Dependency-guided replanning.}
Each update creates a new evidence or message version. The framework invalidates candidate artifacts, decision signatures, and question values that depended on the superseded version, then reruns affected roles in dependency order. Independent evidence and unrelated artifacts remain available. Replies introducing new interpretations also trigger candidate regeneration, allowing the plan to incorporate information beyond its earlier alternatives.

After replanning, the controller refreshes information questions and fallback alternatives, removes resolved questions, and reevaluates the remaining options using the updated state and budgets. The loop continues until a complete response is supported, no feasible clarification has positive estimated net value, or a shared budget is exhausted. The assistant then completes verification and repair to produce the final response.
\FloatBarrier
\section{Experimental Evaluation}
\label{sec:experiments}

We conduct one main experiment on HoosierHelp and two supplementary experiments on NGQA and a FoodGo-based transfer-site selection task. We then examine clarification policies and system components through two controlled ablation studies and assess controller hyperparameter sensitivity.

\begin{table*}[t]
\centering
\caption{HoosierHelp main results. Bold and underlined means mark the best and second-best quality scores.}
\label{tab:main-hoosierhelp}
\begingroup
\normalsize
\setlength{\tabcolsep}{3pt}
\renewcommand{\arraystretch}{1.14}
\begin{tabular*}{\textwidth}{@{\extracolsep{\fill}}ll*{4}{c}@{}}
\toprule
Category & System & Success $\uparrow$ & Tool EM $\uparrow$ & Resource EM $\uparrow$ & Tokens (k/episode) $\downarrow$ \\
\midrule
\multirow{3}{*}{\shortstack[l]{\textit{Single-agent}\\\textit{methods}}} & Qwen3-8B: free questioning & $0.392\pm 0.016$ & $0.407\pm 0.013$ & $0.623\pm 0.026$ & $15.0\pm 0.15$ \\
 & GPT-4.1: free questioning & $0.454\pm 0.025$ & $0.489\pm 0.027$ & $0.653\pm 0.029$ & $17.1\pm 0.5$ \\
 & gpt-oss-120b: free questioning & $0.389\pm 0.006$ & $0.406\pm 0.002$ & $0.629\pm 0.011$ & $10.2\pm 1.5$ \\
\midrule
\multirow{3}{*}{\shortstack[l]{\textit{Multi-agent}\\\textit{coordination}\\\textit{baselines}}} & Our MAS backbone & $0.336\pm 0.002$ & $0.350\pm 0.003$ & $0.510\pm 0.011$ & $17.1\pm 0.1$ \\
 & Majority vote & $0.350\pm 0.010$ & $0.364\pm 0.008$ & $0.525\pm 0.009$ & $35.4\pm 0.2$ \\
 & Text-confidence handoff & $0.324\pm 0.004$ & $0.343\pm 0.005$ & $0.509\pm 0.008$ & $38.8\pm 7.13$ \\
\midrule
\multirow{4}{*}{\shortstack[l]{\textit{Clarification}\\\textit{methods}}} & ProCoT & $0.539\pm 0.008$ & $0.553\pm 0.002$ & $0.678\pm 0.010$ & $30.2\pm 0.1$ \\
 & ATD & $0.533\pm 0.003$ & $0.566\pm 0.008$ & $\underline{0.708}\pm 0.008$ & $30.2\pm 0.2$ \\
 & SAGE-Clarify & $0.539\pm 0.007$ & $\underline{0.581}\pm 0.009$ & $0.706\pm 0.011$ & $29.0\pm 0.1$ \\
 & LLM-VoI & $\underline{0.541}\pm 0.015$ & $0.574\pm 0.015$ & $0.696\pm 0.010$ & $29.7\pm 0.2$ \\
\midrule
 & Ours & $\mathbf{0.578}\pm 0.008$ & $\mathbf{0.599}\pm 0.008$ & $\mathbf{0.721}\pm 0.009$ & $28.2\pm 0.2$ \\
\bottomrule
\end{tabular*}
\endgroup
\end{table*}

\subsection{Experimental Setting}
\label{sec:experimental-setting}

\paragraph{Backbone.}
We use Qwen3-8B~\citep{qwen3technicalreport} for our assistant and all baselines in the main experiment, two supplementary experiments, and both ablations, except the GPT-4.1~\citep{openai2025gpt41} and gpt-oss-120b~\citep{openai2025gptoss120bgptoss20bmodel} single-agent baselines identified in the main results table. Our assistant freezes the backbone weights; candidate generation, planning, clarification, and response revision occur entirely at inference time.

\paragraph{Benchmarks and evaluation construction.}
\textit{HoosierHelp}~\citep{li2026hoosierhelp} evaluates conversational social-service navigation with real service records, structured search tools, and simulated users, covering composite needs, constraint fallback, and non-ideal interactions. We evaluate $240$ instances per run under the official interaction and evaluation protocol, with GPT-4.1-mini as the user simulator. The assistant interacts with the user, retrieves service information through the provided tools, and recommends resources matching the user's requirements. The official protocol allows questions requesting multiple information items; only our controlled ablations impose atomic questions (Section~\ref{sec:ablation-protocol}), so results under the two protocols are not directly comparable.

\textit{NGQA}~\citep{zhang2025ngqa} evaluates personalized nutritional reasoning from user, food, and nutrient information. Our interactive episodes initially withhold selected profile fields, reveal them through clarification, and score final answers against reference annotations. A $20\%$ food-name hash split yields $2{,}652$ test instances; each of three seeds shuffles this pool and evaluates $300$ instances. We report the benchmark's profile-complexity groups as \textit{sparse}, \textit{standard}, and \textit{complex}, with per-seed counts in Appendix~\ref{app:reporting-details}.

\textit{FoodGo}~\citep{rubinstein2020foodgo_data,rubinstein2020foodgo_report} provides food-rescue pickup-and-delivery data. As no existing benchmark evaluates interactive requirement elicitation for provider-side planning, we construct a transfer-site selection task from these records, with $12$ candidate sites and partially disclosed requirement profiles. The assistant clarifies relevant requirements and selects one or two sites. A deterministic planner generates reference decisions from complete profiles, isolating elicitation of fields that change the site set; success thus measures this skill without confounding it with planning quality. The task evaluates requirement elicitation and site selection rather than the original pickup-and-delivery scheduling problem.

\paragraph{Baselines.}
HoosierHelp baselines include single-agent free questioning with Qwen3-8B, GPT-4.1, and gpt-oss-120b; our MAS backbone; majority vote; text-confidence handoff; and four clarification methods: ProCoT~\citep{deng2023proactive}, ATD~\citep{kobalczyk2025atd}, SAGE-Clarify~\citep{suri2026sage}, and LLM-VoI~\citep{dong2026voi}.
NGQA baselines include Role-chain MAS, nine-way majority vote, LLM-Blender~\citep{jiang2023llm}, HybridLLM~\citep{ding2024hybridllm}, and NG-Router~\citep{shi2026ng}.
FoodGo-based baselines include single-agent direct answering, single-agent free questioning, and the same four clarification methods.

The atomic-question study compares the four clarification methods under the controlled protocol, with ask-on-disagreement, fixed-order questioning, and myopic VOI without interaction cost as atomic controls. Component ablations are named by the removed component.

\paragraph{Metrics and reporting.}
We report the following metrics: for Hoo\-sier\-Help, episode success, Tool EM, Resource EM, and total tokens (thousands per episode); for NGQA, overall and profile-com\-plexity-specific binary accuracy, label F1, and explanation ROUGE-L; for the FoodGo-based task, exact-site-set success against the deterministic planner's reference decision under complete requirements.
HoosierHelp success follows the official end-to-end criterion: an explicit final recommendation plus exact tool-call-set and resource-ID-set matches.
Total tokens sum input and output across all backbone calls per episode, including every controller call for candidate sampling, receiver replay, and simulated replies (Appendix~\ref{app:reporting-details}).
All experiments, including both ablations, use three runs; means are computed after rounding each seed's result to three decimal places.
All tables and error bars report mean $\pm$ standard deviation.
NGQA overall accuracy is computed over the $300$ instances in each run before averaging across runs; ablations also report mean questions per episode.
Score differences are in percentage points (pp).
Complete results and evaluation details appear in Appendices~\ref{app:full-results}--\ref{app:reporting-details}.

\subsection{Main Results}
\label{sec:results}
\label{sec:main-results}
\subsubsection{HoosierHelp}
\label{sec:hoosierhelp-results}

Table~\ref{tab:main-hoosierhelp} compares our assistant with single-agent, multi-agent coordination, and clarification baselines on end-to-end resource navigation under the official HoosierHelp interaction protocol~\citep{li2026hoosierhelp}. \textbf{Our system leads on all three metrics}: $57.8\%$ success, $59.9\%$ Tool EM, and $72.1\%$ Resource EM. It beats the strongest baseline, LLM-VoI, by $3.7$ pp on success while spending $5.1\%$ fewer tokens ($28.2$k per episode), the lowest consumption of any clarification method. Higher accuracy at lower cost rules out the explanation that the gain is bought with extra interaction.


\subsubsection{Supplementary Experiments}
\label{sec:secondary-results}
We further evaluate nutritional question answering and provider-side planning, using Qwen3-8B for all systems.

\begin{figure}[htbp]
    \centering
    \includegraphics[width=\columnwidth]{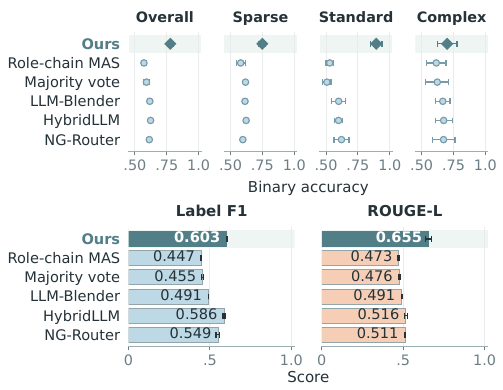}
    \caption{NGQA results. Our assistant achieves the highest accuracy overall and across all profile-complexity groups. It also leads in label F1 and explanation ROUGE-L. Accuracy axes use a restricted range.}
    \Description{Top: overall, sparse, standard, and complex accuracy. Bottom: label F1 and explanation ROUGE-L in the same system order. The full system reports overall accuracy of 0.778, sparse accuracy of 0.750, standard accuracy of 0.894, complex accuracy of 0.700, label F1 of 0.603, and ROUGE-L of 0.655.}
    \label{fig:secondary-evaluations}
    \label{fig:ngqa-quality}
\end{figure}

\paragraph{Nutritional question answering on NGQA}
\label{sec:nutrition-qa}
Figure~\ref{fig:ngqa-quality} shows that \textbf{our system leads on all three NGQA metrics}: binary accuracy $0.778$, label F1 $0.603$, and explanation ROUGE-L $0.655$. The accuracy margin over the best baseline, HybridLLM, is $15.5$ pp, with F1 also ahead ($+0.017$). Accuracy is highest in every profile-complexity group, so the advantage does not depend on profile difficulty. These scores evaluate the complete profile-disclosure pipeline. Table~\ref{tab:ngqa-full} and Appendix~\ref{app:reporting-details} provide full results and sampling details.

\begin{figure}[!htbp]
    \centering
    \includegraphics[width=\columnwidth]{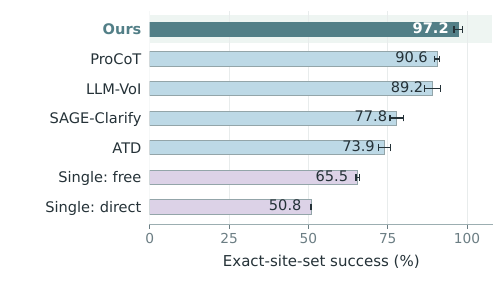}
    \caption{FoodGo transfer-site selection results show that our assistant achieves the highest success rate in recovering the reference site set from partially disclosed provider requirements.}
    \Description{Horizontal bars compare seven systems on provider-side transfer-site selection. Our system achieves the highest success rate at 97.2 percent, followed by ProCoT at 90.6 percent and LLM-VoI at 89.2 percent.}
    \label{fig:foodgo-success}
\end{figure}

\paragraph{Provider-side transfer-site selection.}
\label{sec:foodgo-results}
We construct an interactive transfer-site selection task from FoodGo data~\citep{rubinstein2020foodgo_data,rubinstein2020foodgo_report} to evaluate provider-side planning. Success requires an explicit final recommendation that exactly matches the deterministic planner's site set under the complete provider profile. The reference is produced by the same deterministic planner the agent calls, and this is by design: once the complete profile is known the planner has exactly one answer, so the only way to fail is to leave a decision-changing requirement unasked or to stop before asking it. The task therefore measures requirement elicitation and nothing else. Direct scenarios, half of the set, are solvable without any question and establish a $50\%$ floor that single-agent direct answering reproduces exactly; every point above that floor is earned by eliciting the right fields in the sensitive half. Figure~\ref{fig:foodgo-success} shows that \textbf{our system reaches $97.2\%$ success}, $6.6$ pp above the strongest baseline, ProCoT, under an exact-match criterion that gives no credit for partially correct site sets. Full results and task details appear in Table~\ref{tab:foodgo-full} and Appendix~\ref{app:foodgo-setting}.

\subsection{Ablation Studies}
\label{sec:ablations}

\paragraph{Controlled protocol.}
\label{sec:ablation-protocol}
Both ablations use atomic questions, each requesting one information item or one fallback-acceptance decision, under a cap of eight rounds; policies may stop earlier, and we report actual question counts. The first study removes system components, the second compares clarification policies. This protocol is deliberately different from the main experiment, and numbers are not comparable across the two. The official protocol lets one question request several items, which confounds question selection with question packing; the atomic protocol removes that freedom and charges every item one round. Every policy scores lower under it, and clarification gains less, because less information fits under the cap. The ablations therefore answer one question the main experiment cannot: which component or policy makes the best use of a fixed, item-level budget. The without-clarification variant is the full system with questions disabled under this protocol, not the MAS backbone of Table~\ref{tab:main-hoosierhelp}.

\subsubsection{System Component Ablations}
\label{sec:components}

\paragraph{Without receiver conditioning.}
\label{sec:selection-mechanisms}
The decisive ablation removes receiver conditioning: questions are selected without conditioning on the downstream receiving decisions. In the full system, a premise is relevant exactly when its possible values change a receiver's admissible decisions $\mathcal{A}^{\mathrm{adm}}_j$ or its remaining evidence obligations $\mathcal{O}_j$ (Section~\ref{sec:receiver-relevance}), so a question is worth asking both when the answer can flip a decision and when it supplies the support a fixed decision still lacks. Without this conditioning, the selection no longer sees which premises any receiver depends on.

\begin{figure}[!htbp]
    \centering
    \includegraphics[width=\columnwidth]{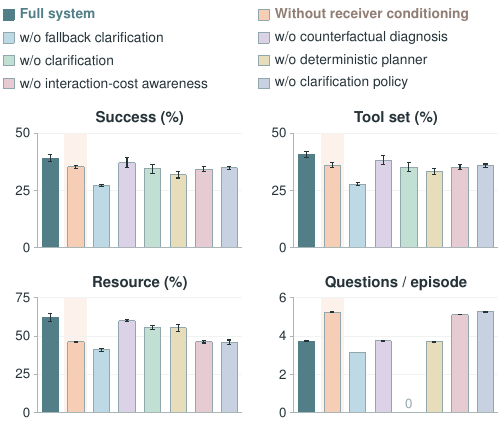}
    \caption{System component ablations show that removing receiver conditioning reduces decision quality while requiring more clarification questions.}
    \Description{Four vertical bar charts compare eight variants on success, tool-call-set matching, resource matching, and questions per episode. Each variant has a distinct color identified in a shared horizontal legend; the full system is teal and without receiver conditioning is peach. Removing receiver conditioning decreases the three quality means from 39.2, 40.7, and 62.3 percent to 35.4, 36.2, and 46.4 percent, while increasing questions from 3.75 to 5.25. Whiskers show standard deviations across three runs}
    \label{fig:component-ablation}
    \label{fig:component-policies}
\end{figure}
\relax\space
Figure~\ref{fig:component-ablation} shows the cost of that change. Success drops from $39.2\%$ to $35.4\%$ and tool-call-set matching from $40.7\%$ to $36.2\%$. Resource matching drops from $62.3\%$ to $46.4\%$ ($15.9$ pp), a direct loss in the correctness of the recommended resource set. At the same time, mean questions rise from $3.75$ to $5.25$ ($+40\%$). \textbf{The ablated system asks $40\%$ more questions and makes worse decisions on every metric.} The pattern is exactly what the mechanism predicts. Without receiver conditioning, the most uncertain premise is asked first whether or not any receiver's decision depends on it, so rounds go to premises whose answers change nothing, and the rounds left for retrieval and repair shrink accordingly. The loss in resource matching is the direct consequence of asking the wrong questions under a fixed cap, and it cannot be recovered by asking more: the ablation already asks more and does worse. Additional questions cannot substitute for the link between missing information and the decisions that consume it; uncertainty alone does not tell the controller which premise is worth the budget. Receiver conditioning is what turns uncertainty representation into useful information gathering under the shared budget.
\paragraph{Other components.}
Every remaining component contributes. Removing fallback clarification is the largest single loss ($-12.0$ pp): information questions whose answers cannot repair a blocked plan still consume rounds, which is why removing all clarification ($34.5\%$) scores \emph{higher} than removing fallback alone ($27.2\%$). Asking without the ability to act on the answer is worse than not asking. Removing the deterministic planner costs $7.2$ pp and counterfactual diagnosis $2.0$ pp. No ablation matches the full system, and none reduces questions without also reducing success.
Every variant that severs the link between questions and receiving decisions converges to the same operating point. Without receiver conditioning, without interaction-cost awareness, without the clarification policy, and the ask-on-disagreement and fixed-order controls of Section~\ref{sec:atomic-quality} all settle at about five questions per episode and resource matching near $46\%$. This is the signature of the removed mechanism, not an artifact. Once selection ignores what the receivers need, the rule used to pick the next premise no longer matters; every such policy spends the cap on decision-irrelevant questions and lands in the same place. The full system is the only configuration that escapes it. Appendix~\ref{app:component-definitions} defines each intervention.

\subsubsection{Atomic-Question Policy Comparison}
\label{sec:atomic-quality}

\begin{figure}[!htbp]
    \centering
    \includegraphics[width=\columnwidth]{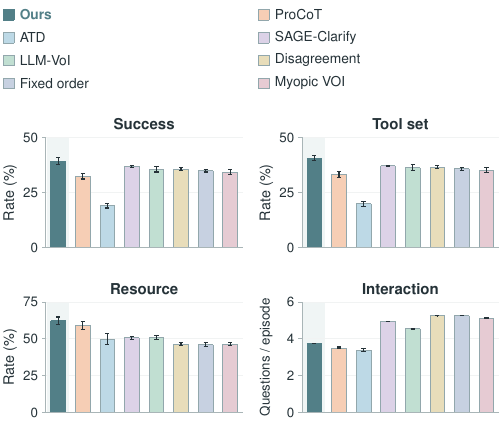}
    \caption{Atomic-question policy comparison shows that our controller achieves the highest decision quality while asking fewer questions than most baselines.}
    \Description{Four panels show success, tool-call-set matching, resource exact matching, and questions per episode for eight policies in the same method order.}
    \label{fig:atomic-clarification}
    \label{fig:atomic-costs}
\end{figure}

\paragraph{Decision quality.}
We compare our controller with ProCoT, ATD, SAGE-Clarify, and LLM-VoI, and with ask-on-disagreement, fixed-order questioning, and myopic VOI without interaction cost, all under the same atomic-question protocol. Figure~\ref{fig:atomic-clarification} shows that \textbf{our controller is the only policy that leads on success, tool matching, and resource matching simultaneously}; every baseline trails on all three. Myopic VOI without interaction cost, the closest decision-theoretic baseline, confirms that the gain does not come from VOI scoring alone but from valuing receiver decisions under a cost-aware budget.

\paragraph{Interaction costs.}
\label{sec:atomic-cost}
We measure mean questions per episode. Figure~\ref{fig:atomic-clarification} shows that our controller asks fewer questions than SAGE-Clarify and LLM-VoI while beating both on success. ProCoT and ATD ask fewer questions but trail on all three quality metrics.

\paragraph{Controller hyperparameter sensitivity.}
Appendix~\ref{app:sensitivity} sweeps $|\Omega|$, $M$, $H$, $c_{\mathrm{item}}$, and $p_{\mathrm{accept}}$ one at a time. Decision quality is flat across the entire grid, within seed variance of the default, while tokens rise monotonically with retained interpretations, branches, and depth. The defaults were fixed outside the evaluation set and sit at the cost knee: no setting buys quality, and larger settings only buy cost. The results do not depend on these choices.

\section{Conclusion}
\label{sec:conclusion}
We presented a conversational food assistant for demand-side resource navigation, nutritional reasoning, and supply-side provision planning. Its receiver-conditioned clarification framework extends single-decision VOI to dependent receiver decisions using shared-premise replay and decision signatures, and jointly values clarification and fallback under a shared budget. Across all tasks, the system outperforms every baseline; under the official HoosierHelp protocol, it also uses fewer tokens than any other clarification method. The ablation supports the central claim: questions should be chosen for their effect on receiver decisions, not premise uncertainty alone. Removing receiver conditioning leads to more questions and worse performance on every decision metric. Jointly evaluating fallback acceptance and its required follow-up questions further connects clarification to the recovery of feasible, evidence-supported plans. These results show the value of grounding multi-agent clarification in the decisions it is meant to improve.
\newpage
\bibliographystyle{ACM-Reference-Format}
\bibliography{references}
\appendix
\section{Complete Numerical Results}
\label{app:full-results}

\begin{table*}[t]
\centering
\caption{NGQA results.}
\label{tab:ngqa-full}
\begingroup
\normalsize
\setlength{\tabcolsep}{3pt}
\renewcommand{\arraystretch}{1.16}
\begin{tabular*}{\textwidth}{@{\extracolsep{\fill}}l*{6}{c}@{}}
\toprule
 & \multicolumn{4}{c}{Accuracy $\uparrow$} & & \\
\cmidrule(lr){2-5}
Method & Overall & Sparse & Standard & Complex & Label F1 $\uparrow$ & ROUGE-L $\uparrow$ \\
\midrule
Role-chain MAS & $0.571\pm 0.020$ & $0.580\pm 0.035$ & $0.529\pm 0.029$ & $0.615\pm 0.076$ & $0.447\pm 0.003$ & $0.473\pm 0.005$ \\
Majority vote (9-way) & $0.591\pm 0.021$ & $0.618\pm 0.014$ & $0.507\pm 0.032$ & $0.623\pm 0.091$ & $0.455\pm 0.006$ & $0.476\pm 0.004$ \\
LLM-Blender & $0.617\pm 0.015$ & $0.614\pm 0.013$ & $0.599\pm 0.055$ & $0.666\pm 0.056$ & $0.491\pm 0.002$ & $0.491\pm 0.002$ \\
HybridLLM & $0.623\pm 0.015$ & $0.623\pm 0.018$ & $0.598\pm 0.033$ & $0.673\pm 0.067$ & $0.586\pm 0.006$ & $0.516\pm 0.008$ \\
NG-Router & $0.613\pm 0.012$ & $0.597\pm 0.013$ & $0.621\pm 0.059$ & $0.672\pm 0.090$ & $0.549\pm 0.012$ & $0.511\pm 0.002$ \\
\midrule
Ours & $\mathbf{0.778}\pm 0.021$ & $\mathbf{0.750}\pm 0.027$ & $\mathbf{0.894}\pm 0.045$ & $\mathbf{0.700}\pm 0.078$ & $\mathbf{0.603}\pm 0.002$ & $\mathbf{0.655}\pm 0.019$ \\
\bottomrule
\end{tabular*}
\endgroup

\end{table*}

\begin{table}[!htbp]
\centering
\caption{Atomic-question policy results. Myopic VOI (no cost) and Fixed order report the same runs as w/o interaction-cost awareness and w/o clarification policy, respectively, in Table~\ref{tab:components-full}.}
\label{tab:atomic-full}
\begingroup
\small
\setlength{\tabcolsep}{1.3pt}
\renewcommand{\arraystretch}{1.16}
\begin{tabular*}{\columnwidth}{@{\extracolsep{\fill}}l*{4}{c}@{}}
\toprule
Policy & Success $\uparrow$ & Tool EM $\uparrow$ & Resource EM $\uparrow$ & Q/ep. $\downarrow$ \\
\midrule
Ours & $\mathbf{0.392}\!\pm\!0.016$ & $\mathbf{0.407}\!\pm\!0.013$ & $\mathbf{0.623}\!\pm\!0.026$ & $3.75\!\pm\!0.02$ \\
\midrule
ProCoT & $0.324\!\pm\!0.012$ & $0.333\!\pm\!0.014$ & $0.592\!\pm\!0.026$ & $3.51\!\pm\!0.05$ \\
ATD & $0.192\!\pm\!0.011$ & $0.200\!\pm\!0.012$ & $0.499\!\pm\!0.035$ & $\mathbf{3.40}\!\pm\!0.09$ \\
SAGE-Clarify & $0.369\!\pm\!0.005$ & $0.372\!\pm\!0.002$ & $0.507\!\pm\!0.009$ & $4.96\!\pm\!0.00$ \\
LLM-VoI & $0.356\!\pm\!0.012$ & $0.364\!\pm\!0.012$ & $0.510\!\pm\!0.013$ & $4.55\!\pm\!0.02$ \\
\midrule
\begin{tabular}[c]{@{}l@{}}Ask on\\disagreement\end{tabular} & $0.357\!\pm\!0.008$ & $0.366\!\pm\!0.008$ & $0.465\!\pm\!0.010$ & $5.26\!\pm\!0.01$ \\
Fixed order & $0.350\!\pm\!0.006$ & $0.359\!\pm\!0.007$ & $0.461\!\pm\!0.014$ & $5.27\!\pm\!0.01$ \\
\begin{tabular}[c]{@{}l@{}}Myopic VOI\\(no cost)\end{tabular} & $0.344\!\pm\!0.011$ & $0.352\!\pm\!0.011$ & $0.464\!\pm\!0.009$ & $5.13\!\pm\!0.01$ \\
\bottomrule
\end{tabular*}
\endgroup
\end{table}

\begin{table}[!htbp]
\centering
\caption{Component ablations (w/o: without).}
\label{tab:components-full}
\begingroup
\small
\setlength{\tabcolsep}{1.3pt}
\renewcommand{\arraystretch}{1.16}
\begin{tabular*}{\columnwidth}{@{\extracolsep{\fill}}l*{4}{c}@{}}
\toprule
Variant & Success $\uparrow$ & Tool EM $\uparrow$ & Resource EM $\uparrow$ & Q/ep. $\downarrow$ \\
\midrule
Full system & $\mathbf{0.392}\!\pm\!0.016$ & $\mathbf{0.407}\!\pm\!0.013$ & $\mathbf{0.623}\!\pm\!0.026$ & $3.75\!\pm\!0.02$ \\
\midrule
\begin{tabular}[c]{@{}l@{}}Without receiver\\conditioning\end{tabular} & $0.354\!\pm\!0.006$ & $0.362\!\pm\!0.010$ & $0.464\!\pm\!0.003$ & $5.25\!\pm\!0.01$ \\
\midrule
\begin{tabular}[c]{@{}l@{}}w/o fallback\\clarification\end{tabular} & $0.272\!\pm\!0.005$ & $0.278\!\pm\!0.007$ & $0.410\!\pm\!0.011$ & $3.17\!\pm\!0.00$ \\
\begin{tabular}[c]{@{}l@{}}w/o counterfactual\\diagnosis\end{tabular} & $0.372\!\pm\!0.021$ & $0.383\!\pm\!0.019$ & $0.601\!\pm\!0.006$ & $3.76\!\pm\!0.03$ \\
w/o clarification & $0.345\!\pm\!0.019$ & $0.353\!\pm\!0.019$ & $0.557\!\pm\!0.012$ & $\mathbf{0.00}\!\pm\!0.00$ \\
\midrule
\begin{tabular}[c]{@{}l@{}}w/o deterministic\\planner\end{tabular} & $0.320\!\pm\!0.015$ & $0.333\!\pm\!0.013$ & $0.554\!\pm\!0.024$ & $3.71\!\pm\!0.03$ \\
\begin{tabular}[c]{@{}l@{}}w/o interaction-cost\\awareness\end{tabular} & $0.344\!\pm\!0.011$ & $0.352\!\pm\!0.011$ & $0.464\!\pm\!0.009$ & $5.13\!\pm\!0.01$ \\
\begin{tabular}[c]{@{}l@{}}w/o clarification\\policy\end{tabular} & $0.350\!\pm\!0.006$ & $0.359\!\pm\!0.007$ & $0.461\!\pm\!0.014$ & $5.27\!\pm\!0.01$ \\
\bottomrule
\end{tabular*}
\endgroup
\end{table}

Tables~\ref{tab:ngqa-full}--\ref{tab:components-full} report mean $\pm$ standard deviation over three runs; quality scores in $[0,1]$, questions per episode. Bold marks the best mean, including ties. Role-chain MAS omits the proposed clarification and recovery mechanisms and is distinct from the without-clarification intervention.
\section{FoodGo Interactive Transfer-Site Selection}
\label{app:foodgo-setting}

\paragraph{Source data and constructed task.}
No public benchmark evaluates interactive requirement elicitation for provider-side food planning. FoodGo~\citep{rubinstein2020foodgo_data,rubinstein2020foodgo_report} is the only released food-rescue dataset we know of with the required pickup, delivery, and volunteer records. Its synthetic test release contains 500 rescue orders (donor/recipient locations, one to three units, perishable flag) and 50 drivers (origin, service interval). Our task adds candidate sites, hidden profiles, clarification, and exact-site-set evaluation; $k$-medoids clustering of available locations yields 12 candidate transfer sites, the complete action space.

\paragraph{Hidden requirements and disclosure.}
An organization representative seeks one or two sites; its six-field profile specifies (1)~donation category: all, perishable only, or bulky only; (2)~objective: shorter donor pickup routes, more pickup locations within a radius, or proximity to recipients; (3)~coverage radius; (4)~operating interval; (5)~whether nearby volunteers are required; (6)~volunteer distance threshold. The opening message states the site count and part of the profile; replies disclose only the requested item.

\paragraph{Tools and interaction limit.}
A deterministic planner scores and ranks all feasible site combinations under supplied constraints; a final-recommendation tool accepts site identifiers and ends the dialogue. Dialogues allow eight agent turns; each question consumes one.

\paragraph{Reference solution and success criterion.}
The same planner enumerates all combinations under the complete hidden profile to obtain the unique optimal set. This deterministic reference introduces no planning error, isolating requirement elicitation: agents succeed by eliciting every decision-changing field, and fail only by omitting one, stopping early, or omitting the final call. Success requires the final-recommendation tool call and exact reference-set matching; missing, extra, or wrong sites fail.

\paragraph{Scenario stratification.}
The 120 scenarios balance site count (one/\allowbreak two) and type (direct/\allowbreak sensitive), 30 per cell. Undisclosed requirements change the optimum in sensitive scenarios; defaults already recover it in direct scenarios. The direct half gives a $50\%$ floor any policy can reach without asking, reproduced by single-agent direct answering; method differences arise from the sensitive half.

\paragraph{Reported results.}
Table~\ref{tab:foodgo-full}: three Qwen3-8B runs over 120 scenarios; highest mean in bold.

\begin{table}[!htbp]
\centering
\caption{FoodGo transfer-site selection results.}
\label{tab:foodgo-full}
\begingroup
\normalsize
\setlength{\tabcolsep}{3pt}
\renewcommand{\arraystretch}{1.16}
\begin{tabular*}{\columnwidth}{@{\extracolsep{\fill}}lc@{}}
\toprule
System & Success $\uparrow$ \\
\midrule
Single agent: direct answer & $0.508\pm 0.000$ \\
Single agent: free questioning & $0.655\pm 0.005$ \\
ProCoT & $0.906\pm 0.008$ \\
ATD & $0.739\pm 0.019$ \\
SAGE-Clarify & $0.778\pm 0.021$ \\
LLM-VoI & $0.892\pm 0.025$ \\
\midrule
Ours & $\mathbf{0.972}\pm 0.013$ \\
\bottomrule
\end{tabular*}
\endgroup
\end{table}
\section{Evaluation Reporting Details}
\label{app:reporting-details}
\label{app:rcc-implementation}

\paragraph{Reporting convention.}
All experiments use three runs; reported means average per-seed results rounded to three decimals; tables and error bars report mean and standard deviation across runs.

\paragraph{HoosierHelp.}
Each run evaluates $240$ instances using Qwen3-8B (except identified GPT-4.1/gpt-oss-120b baselines), with GPT-4.1-mini simulating users under the official setting. Success requires an explicit final recommendation, exact normalized tool-call-set matching, and exact resource-ID-set matching; tool matching covers service, location, schedule, intake, documents, and eligibility~\citep{li2026hoosierhelp}.

\paragraph{Token accounting.}
Reported tokens cover all episode backbone calls, including controller candidate sampling, receiver replay, and simulated replies. Replay reruns only receivers in $\mathcal R_t(q)$ over structured candidates and their evidence, excluding dialogue history; at most six prefiltered questions are evaluated per step; planning and candidate comparison use the deterministic planner. External dialogue dominates cost: each user turn reruns full-history intake, receivers, and retrieval. Scoped replay enables fewer targeted turns: lowest token cost among clarification methods under the official protocol (Table~\ref{tab:main-hoosierhelp}); fewer questions than SAGE-Clarify and LLM-VoI under the atomic protocol (Table~\ref{tab:atomic-full}).

\paragraph{Protocols.}
Official HoosierHelp permits multiple items per question; both ablations permit one item or fallback acceptance per question within eight rounds. We compare no absolute numbers across protocols. Atomic questions isolate selection from packing; fewer items fit under the cap, lowering all policies' scores and compressing clarification gains.

\paragraph{NGQA sampling.}
A food-name hash split reserves $20\%$ for testing: $2{,}652$ instances with \textit{sparse}, \textit{standard}, or \textit{complex} profiles. Each seed shuffles the pool and takes the first $300$; group sizes vary (Table~\ref{tab:ngqa-sampling}), and samples may overlap.

\begin{table}[!htbp]
\centering
\caption{NGQA test-pool and per-seed sample counts.}
\label{tab:ngqa-sampling}
\begingroup
\small
\setlength{\tabcolsep}{3pt}
\renewcommand{\arraystretch}{1.12}
\begin{tabular*}{\columnwidth}{@{\extracolsep{\fill}}lrrrr@{}}
\toprule
Scope & Sparse & Standard & Complex & Total \\
\midrule
Test pool & 1640 & 684 & 328 & 2652 \\
Seed 0 & 194 & 68 & 38 & 300 \\
Seed 1 & 182 & 74 & 44 & 300 \\
Seed 2 & 178 & 82 & 40 & 300 \\
\bottomrule
\end{tabular*}
\endgroup
\end{table}

\paragraph{NGQA aggregation and scoring.}
For group $g$, seed $s$, sample count $n_{s,g}$, and accuracy $a_{s,g}$,
\begin{equation}
a_s=\frac{1}{300}\sum_g n_{s,g}a_{s,g},\qquad
\bar a=\frac{1}{3}\sum_{s=0}^{2}\operatorname{round}_3(a_s).
\end{equation}
Group columns average rounded $a_{s,g}$ across seeds. All systems use Qwen3-8B and staged disclosure: designated profile fields remain hidden until clarification; scoring follows the final answer. Binary accuracy, label F1, and explanation ROUGE-L measure distinct output components. Of $328$ complex-profile instances, $114$ require weighted rules; $214$ also admit simpler rules.

\paragraph{FoodGo and controlled ablations.}
FoodGo uses the $120$ scenarios in Appendix~\ref{app:foodgo-setting}. Atomic question counts reflect actual use, including early stopping; quality-metric definitions are unchanged. Token costs are reported for the official protocol (Table~\ref{tab:main-hoosierhelp}) and controller sensitivity (Table~\ref{tab:sensitivity}).

\section{Component Ablation Definitions}
\label{app:component-definitions}
\label{app:rcc-voi}
Variants in Figure~\ref{fig:component-ablation} and Table~\ref{tab:components-full} name the removed component.
\textit{Without receiver conditioning}: selection is unconditioned on downstream receiving decisions.
\textit{Without fallback clarification}: disables fallback-acceptance questions; retains information questions.
\textit{Without counterfactual diagnosis}: removes counterfactual relaxation diagnosis.
\textit{Without clarification}: disables all questions.
\textit{Without the deterministic planner}: substitutes LLM planning.
\textit{Without interaction-cost awareness}: myopic VOI omits interaction cost, removing cost-aware continuation evaluation and changing lookahead.
\textit{Without the clarification policy}: fixed ordering replaces state-dependent whether/what-to-ask decisions.

\section{Selection Error Bound}
\label{app:selection-bound}
Let $\mathcal C_t=\mathcal Q^{\mathrm{feas}}_t\cup\{\mathrm{proceed}\}$, with $V_t(a)$ and $\widehat V_t(a)$ the expected and estimated net values of action $a$; proceeding has value zero and a question has estimated value $\widehat{\Delta U}_t-\widehat C_t$. Let $a^\star_t=\arg\max_{a}V_t(a)$ and $\widehat a_t=\arg\max_{a}\widehat V_t(a)$. Then
\begin{equation}
\max_{a\in\mathcal C_t}\bigl|\widehat V_t(a)-V_t(a)\bigr|\le\epsilon_t
\;\Longrightarrow\;
V_t(a^\star_t)-V_t(\widehat a_t)\le 2\epsilon_t,
\end{equation}
since $\widehat V_t(\widehat a_t)\ge\widehat V_t(a^\star_t)$ gives $V_t(a^\star_t)\le\widehat V_t(a^\star_t)+\epsilon_t\le\widehat V_t(\widehat a_t)+\epsilon_t\le V_t(\widehat a_t)+2\epsilon_t$.

\section{Ethical Considerations}
\label{app:ethics}
Benchmark and simulated-interaction results do not establish deployment safety or fairness; service or nutritional errors could harm users. Deployment would require verification of current information, protection of sensitive disclosures, user consent for fallback adjustments, and qualified human oversight.
\section{Method Implementation Parameters}
\label{app:method-parameters}

\subsection{System Architecture}
Frozen Qwen3-8B; non-thinking mode; inference only. Workflow: intake $\rightarrow$ three parallel receivers $\rightarrow$ joint planner $\rightarrow$ retriever. Intake maps the full dialogue to candidates with premise identities; receivers read only intake candidates and domain evidence, emitting candidates and decision signatures; the deterministic planner combines signatures and external hard constraints into a jointly feasible plan with consistent fallback adjustments; the retriever maps the planner's retrieval program and results to recommendations.
Receiver roles: navigation---access/\allowbreak eligibility, schedule, logistics; nutrition---profile constraint, food-nutrient, rule applicability; site selection---category/\allowbreak objective, coverage, operations.

Premise fields: \texttt{id}, \texttt{subject}, \texttt{condition}, \texttt{scope}, \texttt{provenance}, \texttt{version}, \texttt{consumers}; forwarding adds consumers without duplicating premises.

\subsection{Candidates and Joint Interpretations}
Sampling: one greedy plus four random draws (temperature $0.7$, top\_p $0.9$); retain $K=3$ candidates including the greedy one. Canonicalization merges equal normalized sets of structured constraint fields; support $w$ is cluster size over total samples. Messages $M_i=\{(a,w,P,F)\}$ contain artifacts, support, provenance including consumed message versions, and scoped assumptions and unresolved requirements. $\Omega$ is the Cartesian product of unresolved-premise candidate values, weighted by products of $w$; retain at most $8$ highest-probability interpretations and renormalize. Each shared premise appears once; each $\omega$ assigns one value used by all dependent receivers during replay.

\subsection{Decisions and Questions}
Signature $\sigma_j(\omega)=(A_j^{\mathrm{adm}}(\omega),O_j(\omega))$: decisions not ruled out and remaining obligations. $Q_{\mathrm{info}}$ targets unresolved premises affecting either, one field per question; unchanged decisions retain questions if $O_j$ is nonempty. $R_t(q)$ comprises receivers reachable from $v(q)$ along dependencies; each scoped premise generates one cross-receiver question. $Q_{\mathrm{fallback}}$ requests acceptance of $D_{\min}$ adjustments. Prefilter $Q_t=Q_{\mathrm{info}}\cup Q_{\mathrm{fallback}}$ by $|R_t(q)|$ and nonempty $O_j$; retain at most $6$ feasible questions per step.

\subsection{Counterfactual Relaxation Diagnosis}
Apply adjustments consistently across affected receivers, holding evidence and external requirements fixed. $D_{\min}=\operatorname{Min}_{\prec}\{\delta\in D_t:r_t(\delta)=1\}$ excludes larger adjustments whenever a smaller one succeeds. Navigation order: time-window relaxation $<$ distance expansion ($\times1.5$, $\times2$) $<$ intake-method relaxation $<$ document-requirement relaxation. Site-selection order: radius expansion ($\times1.5$, $\times2$) $<$ operating-hours relaxation $<$ volunteer-distance requirement removal. Per step: at most $6$ adjustments evaluated, $3$ fallback questions proposed.

\subsection{VOI and Selection}
Each simulated-response branch updates its target premise once, replays affected receivers, activates accepted adjustments, and scores the complete task. Information branches: $M=3$ (top two candidate values and other/unknown); fallback: accept/reject with $p_{\mathrm{accept}}=0.5$. Depth $H=2$: initial question plus at most two follow-ups or retries; one greedy rollout per branch. Simulated user: same backbone, conditioned on sampled $\omega$, answering only the requested field. Select $q^*=\arg\max_{q\in Q_{\mathrm{feas}}}[\Delta U(q)-C(q)]$, where $\Delta U(q)=\mathbb{E}[U(S_\tau)]-U_{\mathrm{proceed}}$ and $C(q)=\mathbb{E}[\sum_\ell c(q_\ell,o_\ell)]$ includes every continuation interaction. Require positive net value; execute only the first question; exclude continuations exceeding remaining budgets from $Q_{\mathrm{feas}}$.

\subsection{Utility and Cost}
$U(S)\in[0,1]$: fraction of active requirements supported by evidence with all $O_j$ cleared; count each premise's improvement once. Inputs: confirmed evidence and retrieval results, never reference answers or hidden profiles. $U_{\mathrm{proceed}}$: same criterion after verification and repair without further questions. Costs: $c_{\mathrm{item}}=0.04$ per information field; $c_{\mathrm{fallback}}=0.06$ per fallback question, including confirmation. Tune only $c_{\mathrm{item}}$, $c_{\mathrm{fallback}}$, and $p_{\mathrm{accept}}$, exclusively outside evaluation; sensitivity in Appendix~\ref{app:sensitivity}.

\subsection{Budgets, Revision, and Termination}
$B_t$: external rounds remaining under the protocol limit; inference caps: $60$ backbone calls per step, $180$ per episode. Response fields: \texttt{subject}, \texttt{condition}, \texttt{scope}, \texttt{provenance}. $S_{t+1}=\operatorname{Revise}(S_t,q_t,o_t)$ updates evidence for information responses and preferences for accepted fallbacks; invalidates candidates, signatures, and question values dependent on old versions; reruns affected receivers and the planner in dependency order. New interpretations regenerate candidates; refreshed question sets discard resolved questions. Stop when a complete answer is supported, no feasible question has positive net value, or a budget is exhausted. Verification/repair: at most $2$ rounds; unsupported recommendations trigger renewed retrieval.

\subsection{Sensitivity to Controller Hyperparameters}
\label{app:sensitivity}
Table~\ref{tab:sensitivity}: one-at-a-time sweeps around defaults ($|\Omega|{=}8$, $M{=}3$, $H{=}2$, $c_{\mathrm{item}}{=}0.04$, $p_{\mathrm{accept}}{=}0.5$), atomic protocol, three seeds per setting. Quality is flat: success $0.383$--$0.404$ except $|\Omega|{=}4$ ($0.376$); resource matching $0.622$--$0.647$. Tokens increase monotonically with interpretations ($24.7$k--$43.4$k), branches ($27.3$k--$29.6$k), and depth ($27.5$k--$30.5$k), without consistent quality gains; defaults sit at the cost knee. Setting $c_{\mathrm{item}}{=}0$ lowers success $0.7$ pp; this scalar change is distinct from removing cost-aware continuation evaluation altogether (Table~\ref{tab:components-full}). For $p_{\mathrm{accept}}\in\{0.3,0.5,0.7\}$, success is $0.404$, $0.392$, $0.396$; retain $0.5$, fixed outside evaluation. Online acceptance estimates from confirmed responses, shrunk toward this prior for sparse histories, are a natural extension; this fixed-prior sweep does not establish their robustness. Mechanism contributions follow from removal (Table~\ref{tab:components-full}), not parameter magnitudes: small $|\Omega|$ still preserves alternatives; shallow $H$ still evaluates continuations.

\begin{table}[!htbp]
\centering
\caption{Controller sensitivity (quality scores in \%).}
\label{tab:sensitivity}
\begingroup
\small
\setlength{\tabcolsep}{2pt}
\begin{tabular*}{\columnwidth}{@{\extracolsep{\fill}}lcccc@{}}
\toprule
Setting & Success $\uparrow$ & Tool set $\uparrow$ & Resource $\uparrow$ & Tokens (k) $\downarrow$ \\
\midrule
Default & $39.2\pm1.6$ & $40.7\pm1.3$ & $62.3\pm2.6$ & $28.2$ \\
\midrule
$|\Omega|{=}4$ & $37.6\pm1.5$ & $39.2\pm1.5$ & $62.2\pm0.6$ & $24.7$ \\
$|\Omega|{=}16$ & $39.0\pm1.6$ & $41.4\pm1.0$ & $63.8\pm1.7$ & $36.7$ \\
$|\Omega|{=}32$ & $39.3\pm0.6$ & $40.8\pm0.6$ & $63.0\pm0.5$ & $43.4$ \\
\addlinespace[2pt]
$M{=}2$ & $38.8\pm1.1$ & $40.1\pm1.1$ & $64.6\pm1.9$ & $27.3$ \\
$M{=}4$ & $39.0\pm2.1$ & $40.6\pm2.7$ & $63.3\pm1.8$ & $28.9$ \\
$M{=}5$ & $39.2\pm0.8$ & $40.6\pm0.9$ & $63.2\pm1.0$ & $29.6$ \\
\addlinespace[2pt]
$H{=}1$ & $38.9\pm1.6$ & $40.6\pm1.7$ & $63.5\pm0.2$ & $27.5$ \\
$H{=}3$ & $39.4\pm0.5$ & $40.9\pm0.5$ & $64.0\pm0.7$ & $30.5$ \\
\addlinespace[2pt]
$c_{\mathrm{item}}{=}0$ & $38.5\pm0.6$ & $40.0\pm0.5$ & $62.3\pm0.6$ & $27.5$ \\
$c_{\mathrm{item}}{=}0.02$ & $38.3\pm0.8$ & $39.8\pm1.2$ & $63.0\pm0.3$ & $27.4$ \\
$c_{\mathrm{item}}{=}0.08$ & $39.2\pm0.7$ & $40.8\pm0.5$ & $63.4\pm1.1$ & $27.4$ \\
\addlinespace[2pt]
$p_{\mathrm{accept}}{=}0.3$ & $40.4\pm1.4$ & $41.4\pm1.3$ & $64.7\pm1.0$ & $28.4$ \\
$p_{\mathrm{accept}}{=}0.7$ & $39.6\pm0.4$ & $41.2\pm0.5$ & $64.1\pm0.8$ & $27.4$ \\
\bottomrule
\end{tabular*}
\endgroup
\end{table}

\section{Food Assistant Interface}

\begin{figure}[!htbp]
    \centering
    \includegraphics[width=\columnwidth]{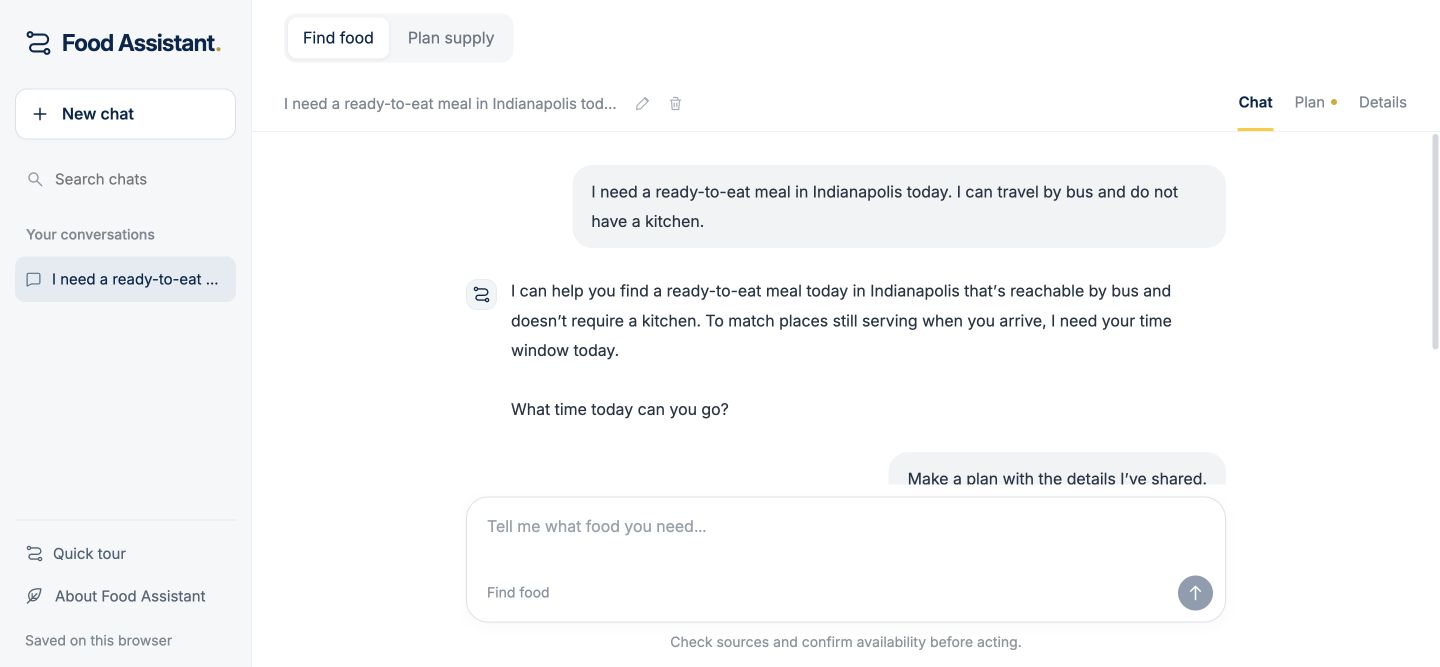}
    \par\smallskip
    \includegraphics[width=\columnwidth]{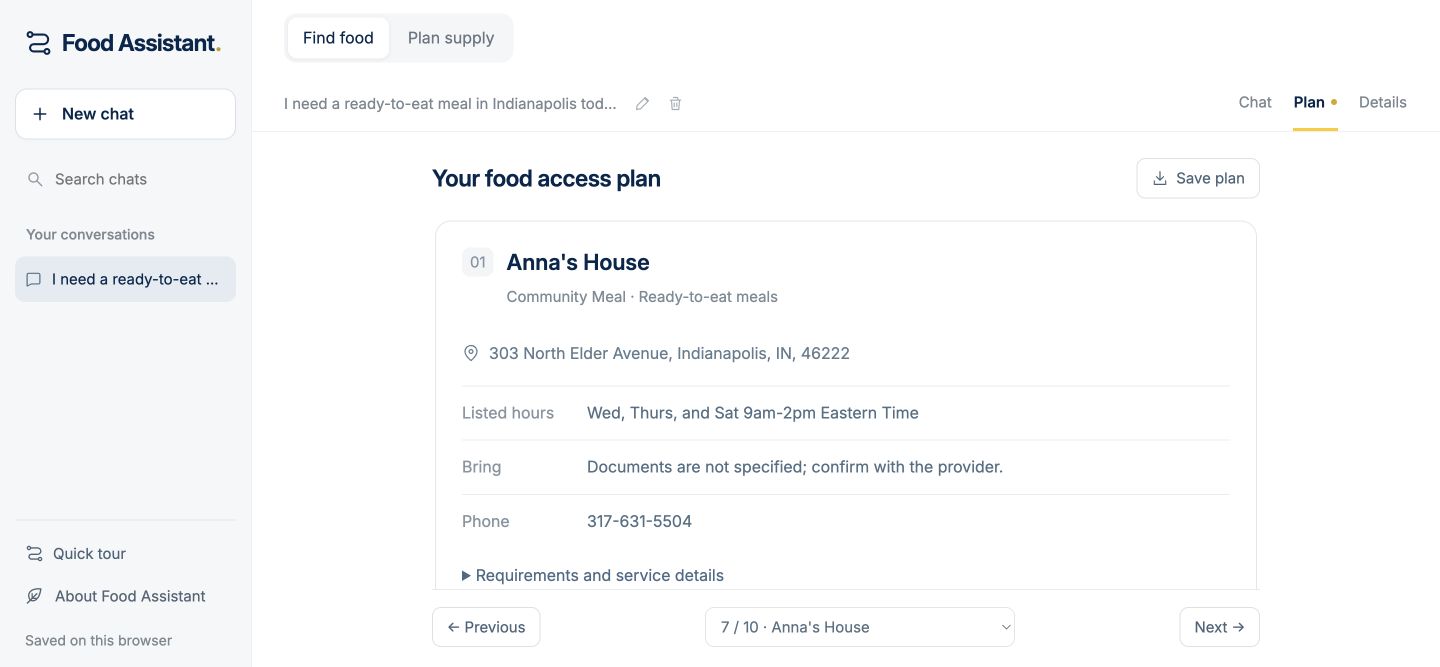}
    \caption{Food Assistant interface: clarification (top) and resource plan (bottom). The plan includes service locations, hours, and access details.}
    \Description{Two screenshots of the Food Assistant. The upper screenshot shows a clarification about the user's time window; the lower screenshot shows a recommended food resource with its service details.}
    \label{fig:food-assistant-interface}
\end{figure}

Figure~\ref{fig:food-assistant-interface} shows the Food Assistant interface. The chat view clarifies user requirements, while the plan view presents recommended resources and service details. Users can switch between both views within the same conversation.

\label{app:food-assistant-interface}

\section{Additional Related Work}
\label{app:additional-related-work}

Large language models (LLMs) have advanced rapidly in recent years \citep{ye2025llms4all, xiong2025deliberate, xiong2026enhancing}. Building on this progress, LLM-driven agents and agentic reinforcement learning have gained increasing attention for their ability to plan, interact with external environments, and solve complex tasks with limited human oversight \citep{zhang2026optimizing, dong2025agentic, dong2026agentic}. In this section, we provide a boarder review of our related work. 

\paragraph{Prior Works in Nutrition Personalization}
With growing awareness of the importance of dietary health, various studies have sought to incorporate health metrics into personalized food recommendation systems to tackle the challenge of food insecurity \cite{zhang2025mopi, zhang2025ngqa, huang2026glen}, aiming at not only provide sufficient but healthy food options. These approaches can be grouped into three primary stages. First, some research emphasizes single indicators like calorie or fat content, as highlighted in works by Ge et al. \cite{ge2015health} and Shirai et al. \cite{shirai2021identifying}, though such metrics often fail to represent the multifaceted nature of a balanced diet. Second, simulated health data has been utilized, as demonstrated by Wang et al. \cite{wang2021market2dish}, but these methods often diverge from real-world data distributions. However, foods deemed healthy by general standards can still negatively affect certain individuals \cite{yue2021overview, zhang2024diet}, highlighting the importance of personalized nutrition in-taking.

\paragraph{Intent uncertainty and clarification timing.}
An assistant can recognize several plausible interpretations of a request before deciding whether to ask the user which one applies. AmbigQA represents the alternatives by pairing plausible answers with disambiguated rewrites~\citep{min2020ambigqa}. CAmbigNQ uses a follow-up question and the user's reply to resolve the ambiguity, while CLAM selectively decides when to ask~\citep{lee2023cambignq,kuhn2022clam}. The timing decision depends on what the uncertainty signal measures. Semantic entropy measures disagreement among sampled answer meanings. SAGE uses sampled answers to calibrate verbal uncertainty, and GCPO calibrates an answer-level uncertainty signal for policy optimization~\citep{kuhn2023semantic,shi2026sage,zhang2026semantic}. IntentSim estimates entropy over plausible user intents and evaluates when to ask under different trade-offs between speed and carefulness~\citep{zhang2025intentsim}. Human question timing provides a separate signal. In the collaborative drawing task annotated by CoDraw-iCR~\citep{madureira2023codraw}, \citet{testoni2024uncertainty} find that human question timing diverges from model uncertainty and that questions guided by model uncertainty improve task success.

\paragraph{Learning from subsequent dialogue and incomplete replies.}
Later turns provide supervision for learning which questions help an assistant proceed. STaR-GATE trains question asking in dialogues with simulated users whose preferences are hidden, using oracle-generated task responses as a reference~\citep{andukuri2024stargate}. Modeling Future Conversation Turns assigns preference labels based on the outcomes of simulated continuations~\citep{zhang2025future}. Even after the user replies, the assistant may still lack a fact needed to act. Not Every User Says It All simulates incomplete replies by masking sentences from user turns. Its StaGPO method assigns credit to individual turns by tracking what the task requires, what remains missing, and how the agent is filling the gap~\citep{bao2026not}. Drift-Bench tests whether agents recover from input faults such as missing parameters, false presuppositions, and ambiguous wording before execution~\citep{bao2026drift}. Food4All traces failures further downstream, separating failures to translate user requirements into search parameters from failures to carry valid search results into the final referral~\citep{li2025food4all}.

\paragraph{Information gain versus decision value.}
A question can narrow the set of plausible interpretations while leaving the preferred action unchanged. Active Task Disambiguation selects questions by their expected information gain over viable solutions~\citep{kobalczyk2025atd}. Qulac evaluates clarification by the quality of the resulting document retrieval~\citep{aliannejadi2019clarifying}. Expected-value question ranking scores the usefulness of anticipated answers, and inference-time value of information weighs the expected utility gain against the cost of asking~\citep{rao2018evpi,dong2026voi}. Decision-region determination formalizes when an agent has enough information to act. Testing can stop once a single decision succeeds under every hypothesis still consistent with the observations~\citep{javdani2014decision}. In a multi-receiver system, this stopping condition can hold for one decision while another still requires clarification. For a food referral, for example, confirming whether the user lives within a pantry's service area can change the eligibility decision while leaving the nutritional assessment unchanged.

\paragraph{Choosing between assistance and additional computation.}
Once a gap matters to a decision, its source determines which intervention can close it. KnowNo uses conformal sets over candidate robot actions to decide when to request assistance. Its sequence-level task-completion guarantee depends on its calibration and scenario-distribution assumptions and on receiving correct assistance~\citep{ren2023knowno}. When the system needs more capable reasoning, HybridLLM routes requests by predicted difficulty and target quality, and AgentRouter learns allocations from agent performance~\citep{ding2024hybridllm,zhang2026agentrouter}. Interpreting a routing score as a probability of correctness requires separate calibration. When the gap concerns external evidence, NG-Router couples routing with evidence retrieval, while CiteAudit verifies the existence and metadata of scientific references~\citep{shi2026ng,shi2026citeaudit}. When the decision turns on an undisclosed constraint known only to the user, a targeted question can supply the missing information.

\end{document}